\documentclass[conference]{IEEEtran}
\IEEEoverridecommandlockouts

\usepackage{subcaption}
\usepackage{multirow}
\usepackage[table,xcdraw]{xcolor}
\usepackage{cite}
\usepackage{amsmath,amssymb,amsfonts}
\usepackage{adjustbox}
\usepackage{algorithmic}
\usepackage{graphicx}
\usepackage{textcomp}
\usepackage{algorithm}
\usepackage{xcolor}
\usepackage{bbding}
\usepackage{geometry}

\def\BibTeX{{\rm B\kern-.05em{\sc i\kern-.025em b}\kern-.08em
    T\kern-.1667em\lower.7ex\hbox{E}\kern-.125emX}}

\newcommand{\ols}[1]{\mskip.5\thinmuskip\overline{\mskip-.5\thinmuskip {#1} \mskip-.5\thinmuskip}\mskip.5\thinmuskip} 
\newcommand{\olsi}[1]{\,\overline{\!{#1}}} 
\makeatletter
\newcommand\closure[1]{
  \tctestifnum{\count@stringtoks{#1}>1} 
  {\ols{#1}} 
  {\olsi{#1}} 
}
\long\def\count@stringtoks#1{\tc@earg\count@toks{\string#1}}
\long\def\count@toks#1{\the\numexpr-1\count@@toks#1.\tc@endcnt}
\long\def\count@@toks#1#2\tc@endcnt{+1\tc@ifempty{#2}{\relax}{\count@@toks#2\tc@endcnt}}
\def\tc@ifempty#1{\tc@testxifx{\expandafter\relax\detokenize{#1}\relax}}
\long\def\tc@earg#1#2{\expandafter#1\expandafter{#2}}
\long\def\tctestifnum#1{\tctestifcon{\ifnum#1\relax}}
\long\def\tctestifcon#1{#1\expandafter\tc@exfirst\else\expandafter\tc@exsecond\fi}
\long\def\tc@testxifx{\tc@earg\tctestifx}
\long\def\tctestifx#1{\tctestifcon{\ifx#1}}
\long\def\tc@exfirst#1#2{#1}
\long\def\tc@exsecond#1#2{#2}
\makeatother

\begin{document}

\title{\textit{L2M-Calib}: One-key Calibration Method for LiDAR and Multiple Magnetic Sensors
}

\author{Qiyang Lyu, Wei Wang, Zhenyu Wu$^{*}$, Hongming Shen, Huiqin Zhou, \\ and Danwei Wang,~\IEEEmembership{Life Fellow,~IEEE}
\thanks{This research is supported by the National Research Foundation, Singapore, under the NRF Medium Sized Centre scheme (CARTIN). Any opinions, findings and conclusions or recommendations expressed in this material are those of the author(s) and do not reflect the views of National Research Foundation, Singapore.}
\thanks{All authors are with the School of Electrical and Electronic Engineering, Nanyang Technological University, Singapore
		{\{zhenyu002\}@e.ntu.edu.sg}}%
    \thanks{*Corresponding author}
}

\maketitle

\begin{abstract}
Multimodal sensor fusion enables robust environmental perception by leveraging complementary information from heterogeneous sensing modalities. However, accurate calibration is a critical prerequisite for effective fusion. This paper proposes a novel one-key calibration framework named \textit{L2M-Calib} for a fused magnetic-LiDAR system, jointly estimating the extrinsic transformation between the two kinds of sensors and the intrinsic distortion parameters of the magnetic sensors. Magnetic sensors capture ambient magnetic field (AMF) patterns, which are invariant to geometry, texture, illumination, and weather, making them suitable for challenging environments. Nonetheless, the integration of magnetic sensing into multimodal systems remains underexplored due to the absence of effective calibration techniques. To address this, we optimize extrinsic parameters using an iterative Gauss-Newton scheme, coupled with the intrinsic calibration as a weighted ridge-regularized total least squares (w-RRTLS) problem, ensuring robustness against measurement noise and ill-conditioned data. Extensive evaluations on both simulated datasets and real-world experiments, including AGV-mounted sensor configurations, demonstrate that our method achieves high calibration accuracy and robustness under various environmental and operational conditions.

\end{abstract}
\section{Introduction}
\begin{figure}
	\centering
	\includegraphics[width=0.9\linewidth]{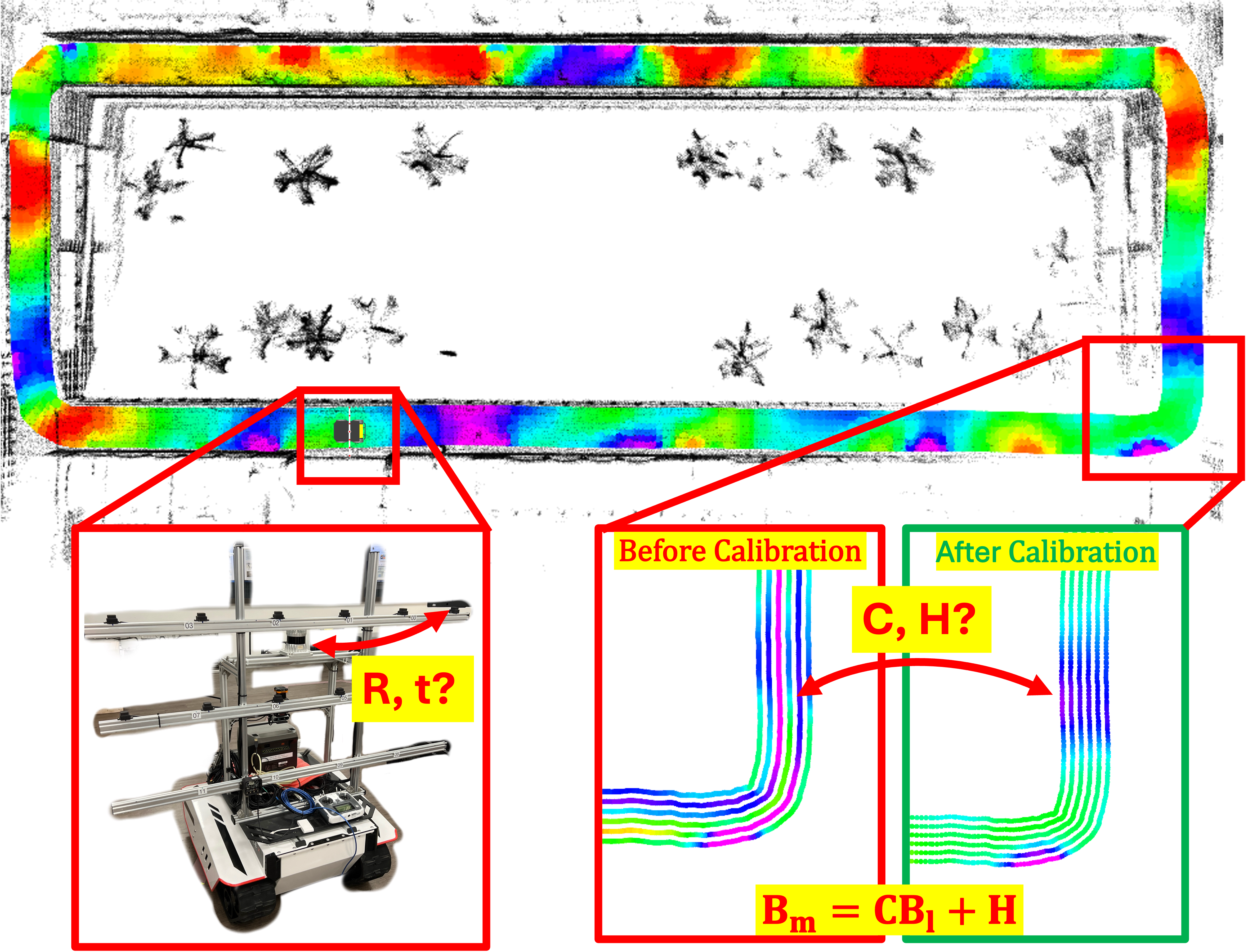}
	\caption{Fusing magnetic sensor with LiDAR requires both the extrinsic parameters of $\mathbf{R, t}$, and the intrinsic parameters of $\mathbf{C, H}$ describing the distortion of magnetic sensors' readings. $\mathbf{B}_\mathrm{m}$ and $\mathbf{B}_\mathrm{l}$ refers to the magnetic readings under magnetic and LiDAR sensor frame, respectively. }
	\label{fig:intro}
	\vspace{-1.5em}
\end{figure}

Sensor fusion, which incorporates multimodal information into a unified perception system, has become an essential component in the development of autonomous ground vehicles (AGVs) 
\cite{itsc_fusion, its_fusion_review}
. Fused sensor data enhances the system's ability to make informed decisions and provides robust performance in tasks such as simultaneous localization and mapping (SLAM) \cite{fuse_sensor_slam}, planning \cite{fuse_sensor_planning}, and navigation \cite{fuse_sensor_navigation}. Existing sensor fusion approaches primarily fall into two categories: relying solely on onboard sensors (e.g., LiDAR, camera, thermal, radar) \cite{garrote2019absolute,Deng_2024_CVPR}, or integrating infrastructure-based beacons (e.g., UWB, Wi-Fi, Bluetooth, RFID) \cite{nazemzadeh2017indoor,greenberg2020dynamic,tzitzis2021real,roos2021mobile}. However, onboard sensor-only solutions can suffer from degraded or even failed performance in environments with ambiguous features. Meanwhile, beacon-based approaches often incur substantial deployment and maintenance costs.

Ambient magnetic field (AMF) information offers a promising alternative for sensor fusion. AMF is an infrastructure-free naturally existing field, distorted locally by nearby ferromagnetic materials, resulting in location-specific features such as vector direction and intensity \cite{wu2020infrastructure}. As ferromagnetic materials are typically built into environments and are not easily altered, the AMF signature remains relatively stable and informative even in visually repetitive or featureless environments like warehouses, tunnels, and seaports. This makes AMF especially advantageous under challenging conditions. Recent studies have increasingly focused on magnetic-field-guided perception \cite{su2020improved, abosekeen2019improving} and demonstrated successful applications in AGV localization tasks \cite{idf_mfl, wu_magloc, wu2024mglt}.

LiDAR, on the other hand, has already been widely adopted in both industry and researches for autonomous systems \cite{fast_lio2, lego_loam, cte_mlo}. LiDAR-based solutions consistently demonstrate high accuracy and reliability. Moreover, cross-sensor calibration methods have been developed for fusing LiDAR with various other sensors, such as LiDAR-camera \cite{velo2cam}, LiDAR-GPS \cite{vision_lidar_gps}, and LiDAR-IMU \cite{lidar_imu} systems.

Given that magnetic sensors provide complementary information to LiDAR and offer unique advantages over other existing sensors, integrating them into a common system can enhance perception robustness. However, aligning their data streams poses significant challenges. In same-modality fusion, data often overlaps spatially, enabling straightforward correspondence establishment. In contrast, the magnetic-LiDAR system involves heterogeneous data modalities without overlapping fields of view. Moreover, raw magnetic sensor measurements are frequently distorted by ferromagnetic elements on the vehicle itself, causing dynamic environmental changes\cite{tollesCompensationAircraftMagnetic1954}. Although these distortions can be compensated through intrinsic calibration after sensor installation, pathological challenges arise when they are mounted on AGVs due to its limited maneuverability
\cite{ningCompensationTechnologyVehicle2024}
.

To address these challenges, we propose a target-free, one-key extrinsic calibration method for magnetic-LiDAR systems, coupled with an intrinsic calibration method to correct carrier-induced magnetic distortions. The main contributions of this work are summarized as follows:

\begin{itemize} 
	\item We design a one-key calibration method that simultaneously estimates both the extrinsic and intrinsic parameters for magnetic-LiDAR fusion systems.  
	\item We introduce a weighted optimization strategy specifically tailored for AGV applications to enhance the robustness of magnetic sensor intrinsic calibration.
	\item We propose a two-step optimization with Gauss-Newton iteration to effectively solve the extrinsic parameters between the magnetic and LiDAR systems.  
\end{itemize}

The remainder of this paper is organized as follows: Section \ref{sec: section2} reviews related work. Section \ref{sec: section3} details the proposed calibration methodology. Section \ref{sec: section4} presents experimental results and detailed analysis. Finally, Section \ref{sec: section5} concludes the paper.

\section{Related Works}\label{sec: section2}
Due to the absence of prior research on extrinsic calibration between magnetic sensors and LiDAR, we first provide a brief review of existing calibration techniques for LiDAR with other sensors. Subsequently, we discuss the influence of intrinsic distortions in magnetic sensors on perception systems and corresponding calibration methods.

\subsection{Calibration of LiDAR to other Sensors}
The calibration of LiDAR to other sensors has been extensively studied. These methods can be broadly categorized into two groups. 

\subsubsection{Calibration via Target-based Methods} In target-based calibration, both sensors operate in similar modalities, allowing them to observe the same physical target. A representative example is the work of Zhou et al. \cite{checkerboard_l2c}, who proposed using a checkerboard for extrinsic calibration of the LiDAR-camera system. Subsequently, Jorge et al. \cite{velo2cam} introduced a four-hole board that improved detection robustness. Zhang et al. \cite{Zhang_thermalcalib} further extended this concept for thermal camera-LiDAR calibration. In 2020, Tóth et al. \cite{sphere_l2c} proposed using a spherical target, which was later adapted by Zhang et al. \cite{l2l_calib} for long-baseline LiDAR-to-LiDAR calibration.

 \subsubsection{Calibration via Trajectory-based Methods.} When two sensors cannot directly observe the same target, calibration methods based on path or motion can be employed. For instance, Lv et al. \cite{aware_calib_l2i} utilized continuous-time trajectories for calibrating LiDAR and IMU systems. Kim et al. \cite{targetless_l2i} proposed a targetless calibration method for ground vehicles, leveraging relative poses constrained by Ground Plane Motion (GPM). Similarly, Chen et al. \cite{lidar_gps} applied hand-eye calibration techniques to estimate the extrinsic parameters between LiDAR and GPS systems.

\subsection{Calibration of Magetic Sensors' Distortion}
Most intrinsic calibration methods for magnetic sensors are based on the classic Tolles-Lawson (T-L) model\cite{tollesCompensationAircraftMagnetic1954}, which aims to estimate parameters that characterize measurement distortions. 
\subsubsection{Orientation-independent Methods Using Magnetic Sensors Only} 
Given the linear nature of the T-L model, early approaches commonly applied Singular Value Decomposition (SVD) \cite{vasconcelosGeometricApproachStrapdown2011} or Least Squares (LS) fitting \cite{chiCalibrationTriaxialMagnetometer2019}. More recent studies have sought to improve upon these techniques. For instance, Chen et al. \cite{chenMagneticFieldInterference2021} employed Total Least Squares (TLS) to account for errors in both the observation vector and the data matrix. 
Ning et al.\cite{ningCompensationTechnologyVehicle2024} further enhanced calibration accuracy under pathological conditions by introducing ridge regularization with L-curve based on TLS. 
\subsubsection{Orientation-dependent Methods Using Auxiliary Sensors} Although pure magnetic sensor-based calibration is effective, it often suffers from high computational cost and strict data collection requirements. To address these issues, auxiliary sensors are incorporated to improve calibration efficiency. For example, Kok et al. \cite{kokMagnetometerCalibrationUsing2016} utilized inertial sensors to refine orientation estimates and correct axis misalignment. Wu et al. \cite{wuDynamicMagnetometerCalibration2018} proposed a dynamic calibration method to minimize misalignment between sensors. Han et al. \cite{hanExtendedKalmanFilterBased2017} adopted an Extended Kalman Filter to fuse gyroscope data, aligning the magnetometer’s rotation with device motion. Similarly, Andel et al. \cite{andelGNSSBasedLowCost2022} used GNSS-based motion vectors to calibrate magnetometers.

Existing research faces several limitations. First, many methods assume that magnetic sensors can provide sufficient excitation, which fails on large or heavy platforms like AGVs with restricted motion, leading to poor data quality and unstable solutions. Second, the assumption of a spatially uniform magnetic field is invalid in environments like warehouses or tunnels, where ferromagnetic interference distorts the field, rendering single ellipsoid fitting ineffective. Third, filter-based approaches often require reliable initialization, which is unavailable in practical scenarios. To address these issues, we propose a new calibration method that avoids such assumptions and remains robust in complex magnetic environments.

\section{The Proposed \textit{L2M-Calib} Method}\label{sec: section3}
\subsection{Extrinsic Model Formulation}
The objective of the extrinsic calibration is to estimate the optimal rotation matrix $\mathbf{R}_\mathrm{m}^\mathrm{l}$ and translation vector $\mathbf{t}_\mathrm{m}^\mathrm{l}$, from the magnetic sensor to the LiDAR frame (\textit{i.e.,} coordinate). This transformation is assumed to be rigid and fixed throughout the calibration process. To obtain reliable excitation data for calibration, the magnetic-LiDAR system is moved along one of various paths (some are illustrated in Fig. \ref{fig: different_paths}) within a prebuilt multi-modal map shown in Fig. \ref{fig:intro}. The prebuilt multi-modal map consists of both a magnetic field map $\mathbb{M}(\mathbf{R}, \mathbf{t})$ and a LiDAR point cloud map $\mathbb{L}(\mathbf{R}, \mathbf{t})$, both referenced in a common global frame. Suppose LiDAR-based localization on $\mathbb{L}(\mathbf{R}, \mathbf{t})$ provides a transformation $(\mathbf{R}_\mathrm{l}^\mathrm{e}, \mathbf{t}_\mathrm{l}^\mathrm{e})$ from the LiDAR frame to the global map frame, given this transformation, the corresponding magnetic sensor pose in the map frame can be derived via: $\mathbf{R}_* = \mathbf{R}_\mathrm{l}^\mathrm{e}\mathbf{R}_\mathrm{m}^\mathrm{l}, \mathbf{t}_* = \mathbf{R}_\mathrm{l}^\mathrm{e}\mathbf{t}_\mathrm{m}^\mathrm{l}+\mathbf{t}\mathrm{_l^e}$. 

To simplify the problem, we first consider the case where the rotation between the LiDAR and magnetic sensor frames is identity ($\mathbf{R}_\mathrm{m}^\mathrm{l} = \mathbf{I_3}$) and the intrinsic distortion of the magnetic sensor is negligible. Under this assumption, the reference magnetic map readings under the magnetic sensor $^\mathrm{ref}\mathbf{B}\mathrm{_m}$ and LiDAR frame $^\mathrm{ref}\mathbf{B}\mathrm{_l}$ would be equal:
$^\mathrm{ref}\mathbf{B}\mathrm{_m} = \mathbb{M}(\mathbf{R}_\mathrm{*}, \mathbf{t}_\mathrm{*}) = {(\mathbf{R}_\mathrm{m}^\mathrm{l})}^{-1}{^\mathrm{ref}}\mathbf{B}\mathrm{_l} =  {(\mathbf{R}_\mathrm{m}^\mathrm{l})}^{-1}\mathbb{M}(\mathbf{R}_\mathrm{l}^\mathrm{e}, \mathbf{t}_\mathrm{*}) = \mathbb{M}(\mathbf{R}_\mathrm{l}^\mathrm{e}, \mathbf{R}_\mathrm{l}^\mathrm{e}\mathbf{t}_\mathrm{m}^\mathrm{l}+\mathbf{t}\mathrm{_l^e})$. 
The residual between the reference map value $^\mathrm{ref}\mathbf{B}\mathrm{_m}$ and the measured sensor reading ${^\mathrm{meas}}\mathbf{B}\mathrm{_m}$  under magnetic sensor frame is thus
\begin{equation}\label{eq:extrinsic_loss}
		\mathbf{e}(\mathbf{t}_\mathrm{m}^\mathrm{l}) = \mathbb{M}(\mathbf{R}_\mathrm{l}^\mathrm{e}, \mathbf{R}_\mathrm{l}^\mathrm{e}\mathbf{t}_\mathrm{m}^\mathrm{l}+\mathbf{t}\mathrm{_l^e}) - {^\mathrm{meas}}\mathbf{B}\mathrm{_m}
\end{equation}
and the optimal translation $\mathbf{t}_\mathrm{m}^\mathrm{l}$ is found by minimizing the squared residual: 
\begin{equation}\label{eq:loss_func}
	\mathbf{t}_\mathrm{m}^\mathrm{l}:=\underset{\mathbf{t}_\mathrm{m}^\mathrm{l}}{\arg\min}\|\mathbf{e}(\mathbf{t}_\mathrm{m}^\mathrm{l})\|^2
\end{equation}

\subsection{Prebuilt Multimodal Magnetic-LiDAR Map}\label{sec: mag-lidar-map}
Due to the non-uniformity of the AMF, we adopt our previous sliding Gaussian Process Regression (s-GPR) method \cite{iecon_lyu} to construct the magnetic map \cite{wu2025magmm} alongside a LiDAR map. A mapping sensor suite \cite{wang2023sensor} with known extrinsic calibration traverses the environment to collect magnetic and LiDAR data. A LiDAR-based SLAM framework estimates the trajectory, while the corresponding magnetic readings are associated with these odometry poses, forming magnetic fingerprints. According to \cite{iecon_lyu}, the magnetic field at any spatial location is modeled as a Gaussian Process:
\begin{equation}
	\begin{split}
		f(\mathbf{t}){\,}{\sim}{\,}\mathcal{GP}{({m(\mathbf{t}), k(\mathbf{t}_p, \mathbf{t}_q))}},\ \ \ \mathbf{B}\mathrm{_e^i} = f(\mathbf{t}_i) + \epsilon
	\end{split}
\end{equation}
where $\mathbf{B}\mathrm{_e^i} = \mathbb{M}(\mathbf{I_3}, \mathbf{t}_i)$ is the $i^\mathrm{th}$ sampled magnetic readings under the global map frame. Then, the posterior distribution over $f(\mathbf{t})$ at an arbitrary point $\mathbf{t}_*$ can be predicted
{\small{\begin{equation} \label{eq:11}
			p({f(\mathbf{t}_*)}|{\,}{\mathbf{t}_*},\mathbf{t},\mathbf{B}\mathrm{_e}) = \mathcal{N}{{\big({{{\mathbb{E}}[\mathbf{B}\mathrm{_e^*}]},\operatorname*{var}(\mathbf{B}\mathrm{_e^*})\big)}}}
\end{equation}}}
with:
{\small{\begin{equation} 
			{{\mathbb{E}}[\mathbf{B}_e^*]} = m(\mathbf{t}_{*}) + {\mathbf{k}_*^{\top}}{\big(\mathbf{K} + {\sigma}_n^2{I_n}\big)}^{-1}(\mathbf{B}\mathrm{_e}-m(\mathbf{t}))
\end{equation}}}
{\small{\begin{equation} \label{eq:13}
			\operatorname*{var}(\mathbf{B}_e^*) = k({\mathbf{t}_*},{\mathbf{t}_*})- {\mathbf{k}_*^{\top}}{\big(\mathbf{K} + {\sigma}_n^2{I_n}\big)}^{-1}{\mathbf{k}_*}
\end{equation}}}
where $\mathbb{M}(\mathbf{I_3}, \mathbf{t}_*) = {{\mathbb{E}}[\mathbf{B}_e^*]}$. To improve computational efficiency, the entire map is partitioned into sub-blocks, with each block interpolated using only nearby training samples.

\subsection{Magnetic Sensor Intrinsic Compensation}
In real-world deployments, magnetic sensors are typically mounted on platforms with ferromagnetic components, leading to sensor distortions. Additionally, misalignment may exist between the LiDAR and magnetic axes. To address these issues, we incorporate both intrinsic distortion and rotation $\mathbf{R}_\mathrm{m}^\mathrm{l}$ back into the calibration model.
\subsubsection{Intrinsic Model Formulation}
Following the Tolles-Lawson (T-L) model \cite{tollesCompensationAircraftMagnetic1954}, the distortion sources include: (1) permanent magnetism, (2) soft iron interference, and (3) eddy current effects (neglected in this scenario\cite{liCompensationMethodCarrier2023}). The complete compensation model becomes: 
\begin{equation}
\label{eq:carrier_intrinsic}
^\mathrm{meas}\mathbf{B}_\mathrm{m}=\mathbf{SC}_{NO}{(\mathbf{R}_\mathrm{m}^\mathrm{l})}^{-1}(\mathbf{K}{^\mathrm{ref}}\mathbf{B}_\mathrm{l}+\mathbf{B}_\mathrm{h})+\mathbf{B}_\mathrm{O}
\end{equation}
where $^\mathrm{meas}\mathbf{B}\mathrm{_m}$ and $^\mathrm{ref}\mathbf{B}\mathrm{_l}$ are the magnetic readings in the magnetic sensors' frame and the reference magnetic field in the LiDAR frame; $\mathbf{S}$, $\mathbf{C}_\mathrm{NO}\in\mathbb{R}^{3\times3}$ and $\mathbf{B}_\mathrm{O}\in\mathbb{R}^{3}$ are the scaling of the magnetic sensor readings for each axis, the non-orthogonality among different axes, and the offset of the origin, respectively; $\mathbf{K}\in\mathbb{R}^{3\times3}$ and $\mathbf{B}_\mathrm{h}\in\mathbb{R}^3$ are the soft iron effect led by temporary magnetization and hard iron effect led by the permanent magnetization of ferromagnetic materials. Eq.~\ref{eq:carrier_intrinsic} can be simplified into linear form:
\begin{equation}
\label{eq:simp_carrier_intrinsic}
\mathbf{B}_\mathrm{m} = \mathbf{C}\mathbf{B}_\mathrm{l} + \mathbf{H}\\
\end{equation}
where $\mathbf{B}_\mathrm{m}, \mathbf{B}_\mathrm{l}$ is simplified from $^\mathrm{meas}\mathbf{B}\mathrm{_m}, {^\mathrm{ref}}\mathbf{B}\mathrm{_l}$.  Given $N > 4$ magnetic reading pairs ${(\mathbf{B}_\mathrm{m}^i, \mathbf{B}_\mathrm{l}^i)}_{i=1}^N$, parameters $\mathbf{C}\in\mathbb{R}^{3\times3}$ and $\mathbf{H}\in\mathbb{R}^{3}$ can be estimated via linear regression.

\subsubsection{Linear Regression under Pathological Situation}
To account for noise in both measurements and system matrix, we adopt the Total Least Squares (TLS):
\begin{equation}
    \mathbf{B}\mathrm{_m}+\mathbf{F} = (\mathbf{B}\mathrm{_l}+\mathbf{E})\mathbf{A}^{\top}
\end{equation}
where $\mathbf{A}=[\mathbf{C}~~\mathbf{H}]$, and with objective:
\begin{equation}
    \arg\min_{\mathbf{A,E,F}}\|[\mathbf{E}~~\mathbf{F}]\|^2_\mathrm{F}
\end{equation}
where $\|\cdot\|_\mathrm{F}$ is Frobenius norm. TLS is solved via truncated SVD on the extended matrix $\mathbf{G} = [\mathbf{B}_\mathrm{m} \ \mathbf{B}_\mathrm{l}]$, retaining dominant components. The reconstructed extension matrix after truncation is $\bar{\mathbf{G}}=[\bar{\mathbf{B}}\mathrm{_e}~~\bar{\mathbf{B}}\mathrm{_l}]=\bar{\mathbf{U}}\bar{\mathbf{S}}\bar{\mathbf{V}}^T$.
When sensor motion is constrained (e.g., mounted on grounded vehicles), observation matrices become ill-conditioned. To improve robustness, regularization is introduced\cite{ningCompensationTechnologyVehicle2024}:
\begin{equation}
			\label{eq:opt_rrtls}
			\arg\min_\mathbf{A}\|\bar{\mathbf{B}}\mathrm{_l}\mathbf{A}^{\top}-\bar{\mathbf{B}}\mathrm{_m}\|^2 + \lambda^2\|\bar{\mathbf{B}}\mathrm{_l}\|^2
\end{equation}
where $\lambda$ is the regularization factor.

\subsubsection{Weighted Regression based on Magnetic Map}
In Section \ref{sec: mag-lidar-map}, the magnetic map is built with Gaussian Process. And in Eq. \ref{eq:13}, the variance $\mathrm{var}(\mathbf{B}_\mathrm{e}^*)$ at the interpolated points $\mathbf{t}_*$ can be obtained in addition to the predicted value. 
The variance from the GP map reflects prediction confidence.
\begin{equation}
    \mathbf{W}={\operatorname{diag}(\operatorname{var}(\mathbf{B}\mathrm{_e^*}))}^{-1}
\end{equation}
This leads to a weighted ridge-regularized TLS formulation:
\begin{equation}
	\label{eq:opt_w-rrtls}
	\arg\min_\mathbf{A}\|\mathbf{W}(\bar{\mathbf{B}}\mathrm{_l}\mathbf{A}^{\top}-\bar{\mathbf{B}}\mathrm{_m})\|^2 + \lambda^2\|\bar{\mathbf{B}}\mathrm{_l}\|^2
\end{equation}
with the closed-form solution:
\begin{equation}\label{eq:w-RRTLS}
    \mathbf{A}_\mathrm{w-RRTLS}=\bar{\mathbf{B}}\mathrm{_m}^{\top}\mathbf{W}^{\top}\mathbf{W}\bar{\mathbf{B}}\mathrm{_l}(\bar{\mathbf{B}}\mathrm{_l}^{\top}\mathbf{W}^{\top}\mathbf{W}\bar{\mathbf{B}}\mathrm{_l}+\lambda\mathbf{I})^{-1}
\end{equation}

\subsubsection{Joint Parameters Optimization}
Substituting Eq. \eqref{eq:simp_carrier_intrinsic} into the residual model Eq. \eqref{eq:extrinsic_loss} yields:
\begin{equation}
    \mathbf{e(C, H,} \mathbf{t}\mathrm{_m^l}) = \mathbf{C}\mathbb{M}(\mathbf{R}\mathrm{_l^e}, \mathbf{R}_\mathrm{l}^\mathrm{e}\mathbf{t}_\mathrm{m}^\mathrm{l}+\mathbf{t}\mathrm{_l^e}) + \mathbf{H}- \mathbf{B}_\mathrm{m}
\end{equation}
Note that the solution of the extrinsic parameter $\mathbf{R}\mathrm{_m^l}$ has been implictly included in the parameter $\mathbf{C}$ and can be directly obtained via the closed form solution given a specific $\mathbf{t}\mathrm{_m^l}$ with Eq. \ref{eq:w-RRTLS}. The optimization is then simplified to solely rely on the extrinsic parameter $\mathbf{t}\mathrm{_l^m}$, with the target loss function of Eq. \ref{eq:loss_func}. To find the optimal $\mathbf{t}\mathrm{_l^m}$, Gauss-Newton optimization is utilized. The corresponding Jacobian matrix with reference to $\mathbf{t}\mathrm{_l^m}$ can be calculated as:
\begin{equation}
    \mathbf{J}(\mathbf{t}_\mathrm{m}^\mathrm{l}) = \frac{\partial{\mathbf{e}}}{\partial{\mathbf{t}_\mathrm{m}^\mathrm{l}}} =\mathbf{R}_\mathrm{l}^\mathrm{e}\mathbf{C}\nabla\mathbb{M}(\mathbf{R}\mathrm{_l^e}, \mathbf{R}_\mathrm{l}^\mathrm{e}\mathbf{t}_\mathrm{m}^\mathrm{l}+\mathbf{t}\mathrm{_l^e})
\end{equation}
and the resulting incrementals of $\mathbf{t}_\mathrm{m}^\mathrm{l}$ at each step can be solved using:
\begin{equation}
        \mathbf{J}(\mathbf{t}_\mathrm{m}^\mathrm{l})^{\top}\mathbf{J}(\mathbf{t}_\mathrm{m}^\mathrm{l})\Delta{\mathbf{t}}=-\mathbf{J}(\mathbf{t}_\mathrm{m}^\mathrm{l})\mathbf{e}(\mathbf{t}_\mathrm{m}^\mathrm{l})
\end{equation}
which eventually led to a two-step stochastic optimization problem that enables one-key joint calibration of both intrinsic and extrinsic parameters across multiple magnetic sensors.
\vspace{-1em}
\begin{algorithm}
    \caption{One-key Calibration Algorithm for Both Extrinsic and Intrinsic Parameters}
    \begin{algorithmic}[1]\label{alg: full}
        \STATE $\mathbf{t}_0\gets[0, 0, 0], \mathrm{maxIteration}=100$
        \FOR{$k < \mathrm{maxIteration}$}
            \STATE Project magnetic sensors to map positions: $\mathbf{t}_* = \mathbf{R}_\mathrm{l}^\mathrm{e}\mathbf{t}_\mathrm{m}^\mathrm{l}+\mathbf{t}\mathrm{_l^e}$
            \STATE Get reference magnetic map value: $[{{\mathbb{E}}[\mathbf{B}_e^*]}, {{\mathrm{var}}[\mathbf{B}_e^*]}]$, $\mathbb{M}(\mathbf{I_3}, \mathbf{t}_*)={{\mathbb{E}}[\mathbf{B}_e^*]}$
            \STATE Rotate back to the LiDAR sensor's frame: $\mathbf{B}_\mathrm{l}={(\mathbf{R}_\mathrm{l}^\mathrm{e})}^{-1}\mathbf{B}_\mathrm{e}^* = \mathbb{M}(\mathbf{R}_\mathrm{l}^\mathrm{e}, \mathbf{t}_*)$
            \STATE $\mathbf{X}\gets\mathbf{B}_\mathrm{l}^*, \mathbf{Y}\gets\mathbf{B}_\mathrm{m},\mathbf{W}\gets{\operatorname{diag}(\operatorname{var}(\mathbf{B}\mathrm{_e^*}))}^{-1}$
            \STATE Solve intrinsic parameters: $[\mathbf{C}~~\mathbf{H}] \gets \mathbf{A}_\mathrm{W-RRTLS}$
            \STATE Calculate cost function: $\mathbf{e(C, H,} \mathbf{t}\mathrm{_m^l})$
            \STATE Calculate Increments: $ \mathbf{J}(\mathbf{t})^{\top}\mathbf{J}(\mathbf{t})\Delta{\mathbf{t}}=-\mathbf{J}(\mathbf{t})\mathbf{e}(\mathbf{t})$
            \STATE Update: $\mathbf{t}_\mathrm{m}^\mathrm{l}  = \mathbf{t}_\mathrm{m}^\mathrm{l} +\Delta \mathbf{t}$
        \ENDFOR
    \end{algorithmic}
\end{algorithm}

\section{Experiments and Analysis}\label{sec: section4}
 \begin{figure} 
	\centering 
	\includegraphics[width=0.99\linewidth]{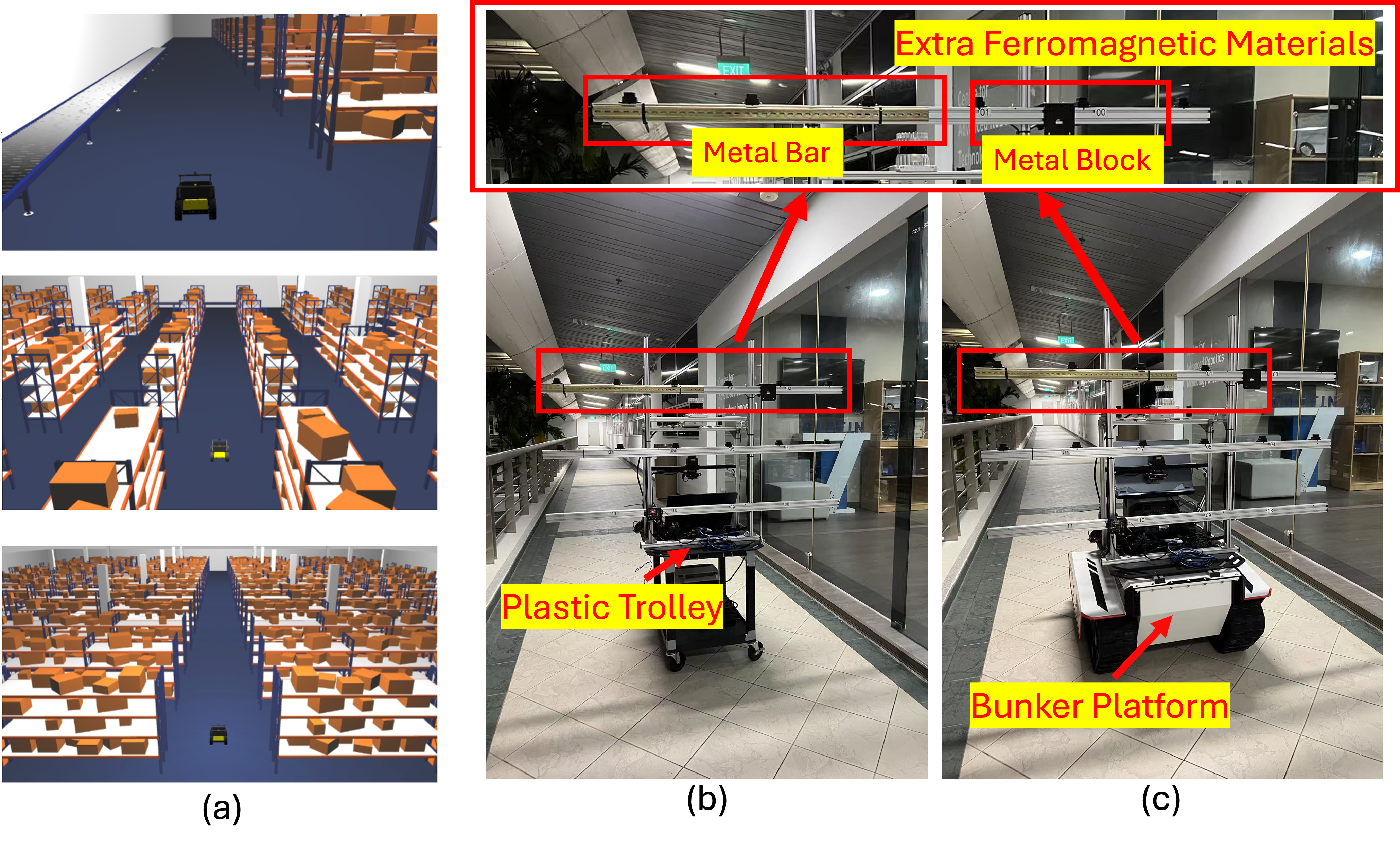} 
	\caption{(a) High-fidelity simulated warehouse environment; (b)(c) The trolley and AGV platform used in real-world experiments.} 
	\label{fig: exp_setup} 
	\vspace{-1.5em}
\end{figure}
\subsection{Evaluation Overview}
To validate the effectiveness of the proposed calibration method, we conduct both simulation-based and real-world experiments. All experiments are executed on a computer equipped with an Intel i9-13200 @ 2GHz CPU and 32 GB RAM. The proposed method is quantitatively assessed in a high-fidelity simulation environment with available ground-truth data and further evaluated for practical applicability and robustness in real-world scenarios using an AGV platform. The following setups are used:
\begin{itemize}
    \item High-fidelity Simulations: The Gazebo simulator is employed to construct a realistic industrial warehouse environment measuring $45\text{m} \times 35\text{m}$, featuring repetitive storage racks and steel-reinforced concrete pillars. A Clearpath Husky A200 AGV is used, with simulated RM3100 magnetic sensors mounted on the frame of the Husky body. Artificial distortions due to the AGV body, along with Gaussian noise, are added to the magnetic data. A simulated OS1-32 LiDAR is mounted centrally on the AGV. Ground-truth odometry information is retrived from the Gazebo simulator directly. The simulated AMF readings are then time-synchronized and projected onto the odometry trajectory.
    \item Real-world Experiments: A plastic trolley platform is deployed to collect AMF data in a corridor environment measuring $51\text{m} \times 17\text{m}$. Twelve pre-calibrated RM3100 magnetic sensors are mounted on the AGV's aluminum frame, forming an ideal magnetic measurement system assumed to be distortion-free. Extra ferromagetnic materials are added later for calibration. Odometry information is derived using the state-of-the-art LiDAR-inertial odometry algorithm FAST-LIO2~\cite{fast_lio2}. The magnetic readings are projected onto the odometry using time synchronization.
\end{itemize}

\subsection{Evaluation Protocol}
\subsubsection{Comparison Baseline}
The proposed \textit{L2M-Calib} method is compared with two state-of-the-art calibration methods to illustrate its effectiveness and advantages. Specifically: 1) OLS: Ordinary Least Squares\cite{chiCalibrationTriaxialMagnetometer2019}; 2) RRTLS: Ridge Regularized Total Least Squares\cite{ningCompensationTechnologyVehicle2024}. In simulation, each method's accuracy is quantified by comparing calibration results against known ground-truth values. In real-world experiments, where ground-truth is unavailable, performance is assessed through statistical comparison between the recovered magnetic readings and a validation magnetic map.
\subsubsection{Evaluation Metrics}\label{sec:eva_met}
Calibration accuracy is evaluated by: 1) Extrinsic Parameter Accuracy: The translation vector $\mathbf{t}_\mathrm{m}\mathrm{l}$ are evaluated via squared Euclidean norms: $e_{\mathbf{t}}=\|\hat{\mathbf{t}}-\mathbf{t}_{\mathrm{gt}}\|_{2}^{2}$; 2) Intrinsic Parameter Accuracy: The distortion matrix $\mathbf{C}$ is evaluated using the Frobenius norm:  $e_\mathbf{C}=\|\hat{\mathbf{C}}-\mathbf{C}\mathrm{_{gt}}\|\mathrm{_F}$ and the bias vector $\mathbf{H}$ is evaluated using squared Euclidean norms: $e_{\mathbf{H}}=\|\hat{\mathbf{H}}-\mathbf{H}_{\mathrm{gt}}\|_{2}^{2}$; 3) Real-world performance: For real-world experiments, where $\mathbf{C}_{\text{gt}}, \mathbf{H}_{\text{gt}}, \mathbf{t}_{\text{gt}}$ are unknown, the accuracy is assessed using the mean squared error between the calibrated magnetic readings $\mathbf{B}_\mathrm{AC}$ and the readings from validation map $\mathbf{B}_\mathrm{MR}$: $e_{\mathbf{B}_\mathrm{AC}} = \mathrm{mean(\|{\mathbf{B}_\mathrm{AC}} - \mathbf{B}_\mathrm{MR}\|_2^2)}$

\subsection{Quantitative Analysis}
\subsubsection{Full Calibration Accuracy}
 \begin{figure} 
	 \centering 
	 \includegraphics[width=0.99\linewidth]{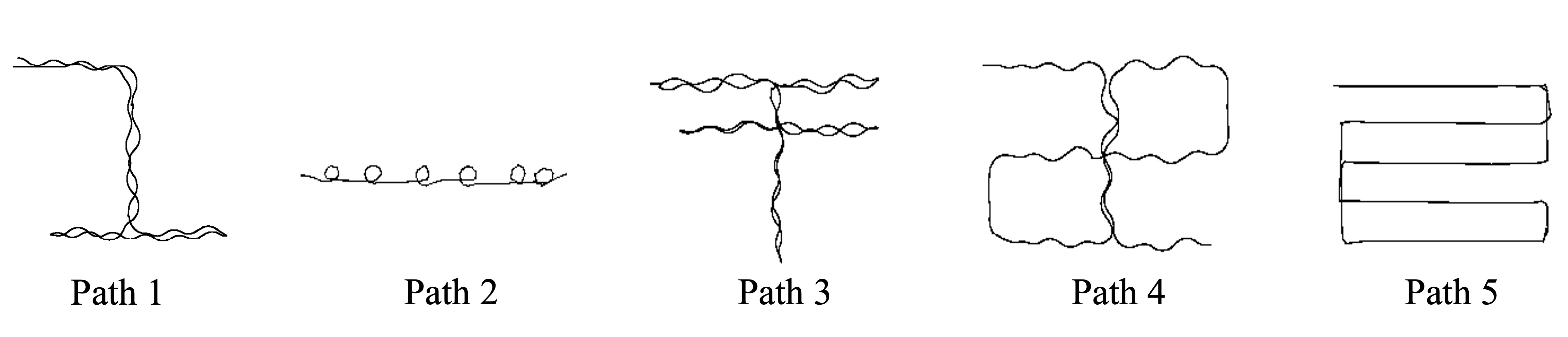} 
	 \caption{Different random paths generated for collecting magnetic sensor's readings for calibration.} 
	 \label{fig: different_paths} 
	 \vspace{-1.0em}
\end{figure}

\begin{table}[!t]
	\caption{Transformation $\mathrm{e}_\mathbf{t}$, distortion $\mathrm{e}_\mathbf{C}$ and bias $\mathrm{e}_\mathbf{H}$ parameter error after two-step calibration under different paths and noise levels.}
	\label{tab: trans_err_simu}
	\centering
	\begin{adjustbox}{width=0.99\linewidth}
		\begin{tabular}{cl|ccc|ccc}
			\hline
			\multicolumn{1}{l}{}         &        & \multicolumn{3}{c|}{Mean ($\mu T$)}                                    & \multicolumn{3}{c}{Std}                                      \\
			&        & {noise=0.1$$} & {noise=0.2} & {noise=0.5} & {noise=0.1} & {noise=0.2} & {noise=0.5} \\ \hline
			\multirow{5}{*}{$\mathbf{e_t}$} & Path 1 & \textbf{0.0018}             & 0.0037             & 0.0058             & 0.0009             & 0.0014             & 0.0030             \\
			& Path 2 & 0.0031             & 0.0029             & 0.0076             & 0.0007             & 0.0016             & 0.0053             \\
			& Path 3 & 0.0039             & 0.0021             & \textbf{0.0034}             & \textbf{0.0005}             & 0.0012             & \textbf{0.0018
			}\\
			& Path 4 & 0.0025             & \textbf{0.0017}             & 0.0048             & 0.0012             & 0.0014             & 0.0039             \\
			& Path 5 & 0.0049             & 0.0045             & 0.0062             & 0.0008             & \textbf{0.0010}             & 0.0020             \\ \hline
			\multirow{5}{*}{$\mathbf{e_C}$} & Path 1 & 0.0018             & 0.0055             & 0.0289             & \textbf{0.0010}             & \textbf{0.0089}             & \textbf{0.0885}             \\
			& Path 2 & 0.0029             & 0.0080             & 0.0427             & 0.0069             & 0.0235             & 0.1331             \\
			& Path 3 & \textbf{0.0016}             & \textbf{0.0047}             & \textbf{0.0276}             & 0.0023             & 0.0136             & 0.0886             \\
			& Path 4 & 0.0028             & 0.0070             & 0.0378             & 0.0066             & 0.0215             & 0.1154             \\
			& Path 5 & 0.0028             & 0.0070             & 0.0368             & 0.0047             & 0.0205             & 0.1218             \\ \hline
			\multirow{5}{*}{$\mathbf{e_H}$} & Path 1 & 0.0751             & 0.1400             & 0.2623             & \textbf{0.0156}             & \textbf{0.0468}             & 0.5669             \\
			& Path 2 & 0.0508             & 0.0979             & 0.5559             & 0.0485             & 0.1658             & 0.9575             \\
			& Path 3 & \textbf{0.0300}             & \textbf{0.0622}             & \textbf{0.2541}             & 0.0164             & 0.0812             & \textbf{0.5643}             \\
			& Path 4 & 0.0398             & 0.0748             & 0.4048             & 0.0378             & 0.1296             & 0.7131             \\
			& Path 5 & 0.0673             & 0.0923             & 0.3307             & 0.0163             & 0.1111             & 0.7460             \\ \hline
		\end{tabular}
	\end{adjustbox}
	\vspace{-1.5em}
\end{table}

In order to show the effectiveness of the overall calibration algorithm, calibration data are collected over five different paths independently shown in Fig. \ref{fig: different_paths} and three different noise levels. At each noise level, fifty random distortion parameters $\mathbf{C, H}$ are generated and added to the ideal simulated magnetic sensor. Then, with each random distortion added, fifty random offsets ranging from $0 - 1m$ are added to the ground truth position as the initial position $\mathbf{t}_0$ for optimization. The mean and standard deviation of the evaluation metrics mentioned in Sec. \ref{sec:eva_met} are calculated. The final results are shown in Table \ref{tab: trans_err_simu}.

It can be seen that the overall calibration accuracy under simulation environments are very high. Even under increased noise, extrinsic errors remain below $1\text{cm}$, and standard deviations are consistently under $1\text{cm}$, given the initial position's scale up to $1m$. The accurate extrinsic parameter also helps ensure the successful calibration of the intrinsic parameter, where the bias errors are smaller than $10nT$.  It also can be observed that the calibration accuracy is independent on the choice of paths, indicating robustness to trajectory variation.

\subsubsection{Extrinsic Calibration Success Rate}
 \begin{figure} 
	\centering 
	\includegraphics[width=0.99\linewidth]{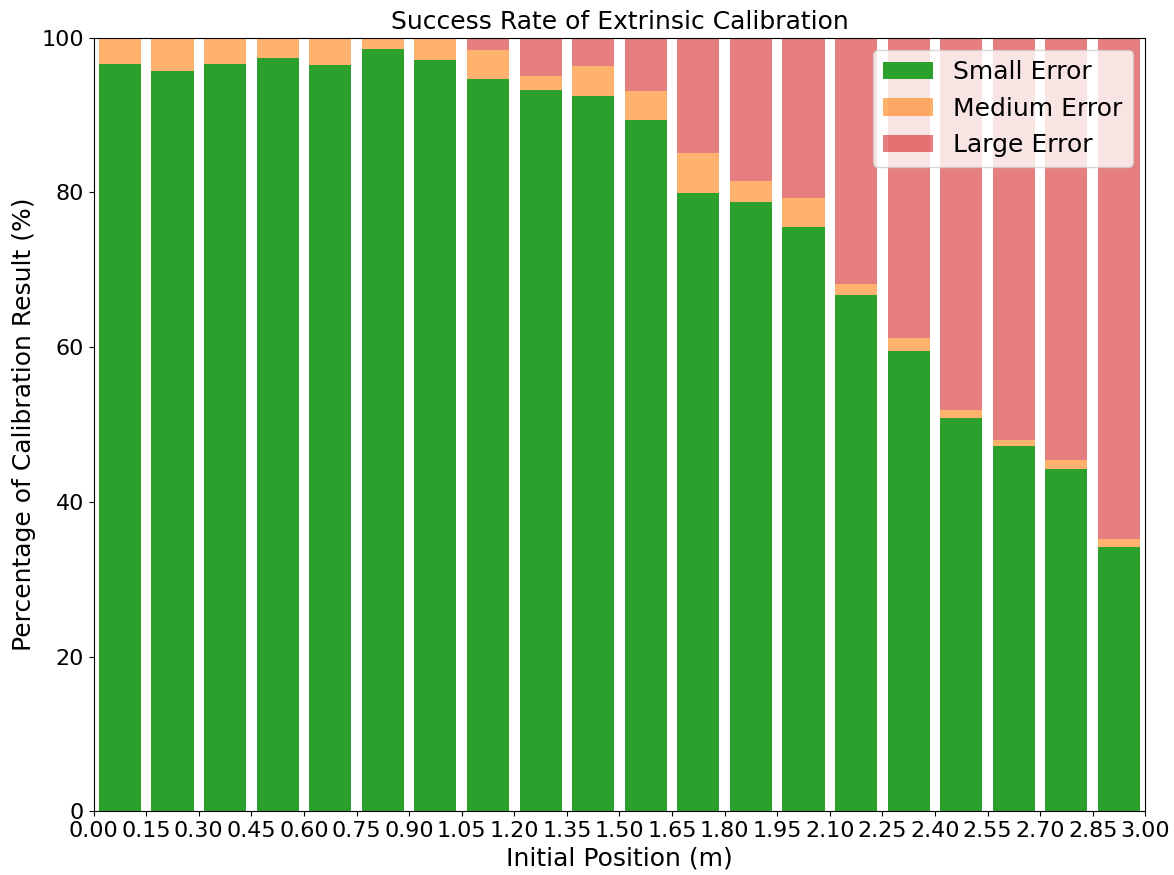} 
	\caption{Success rate of extrinsic parameter calibration under different initial estimation.} 
	\label{fig: success_rate} 
	 \vspace{-2em}
\end{figure} 
To further examine robustness, we evaluate success rate under larger initial offsets. For twenty randomly generated distortion parameters, 100 random initial positions within $0$–$3\text{m}$ are tested. A calibration is considered successful if the final estimate of extrinsic parameter lies within $2\text{cm}$ (small error), within $5\text{cm}$ (medium error), or otherwise marked as failure. Success rates are reported in Fig.~\ref{fig: success_rate}, illustrating that \textit{L2M-Calib} maintains a high success rate within $1.5\text{m}$ of the true initial position, showing a good tolerance to initial guess errors.

\subsubsection{Ablation Study: Effect of Weighting on the Calibration Result}
\begin{figure*}[!t]
	\centering{
		\subfloat[Interpolation result ]{
			\includegraphics[width=0.45\linewidth]{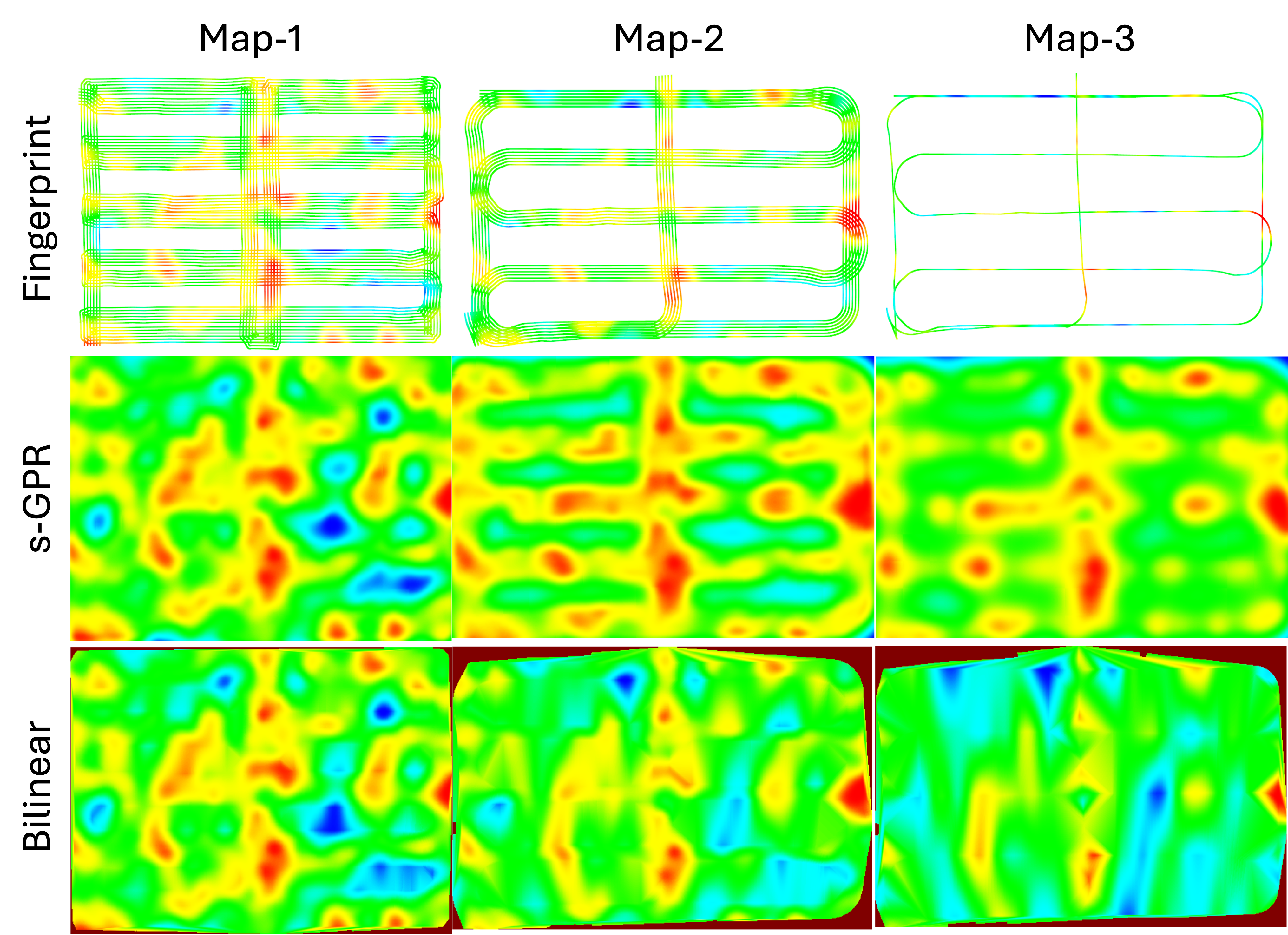}
			\centering
		}
		\subfloat[Calibration result]{
			\includegraphics[width=0.45\linewidth]{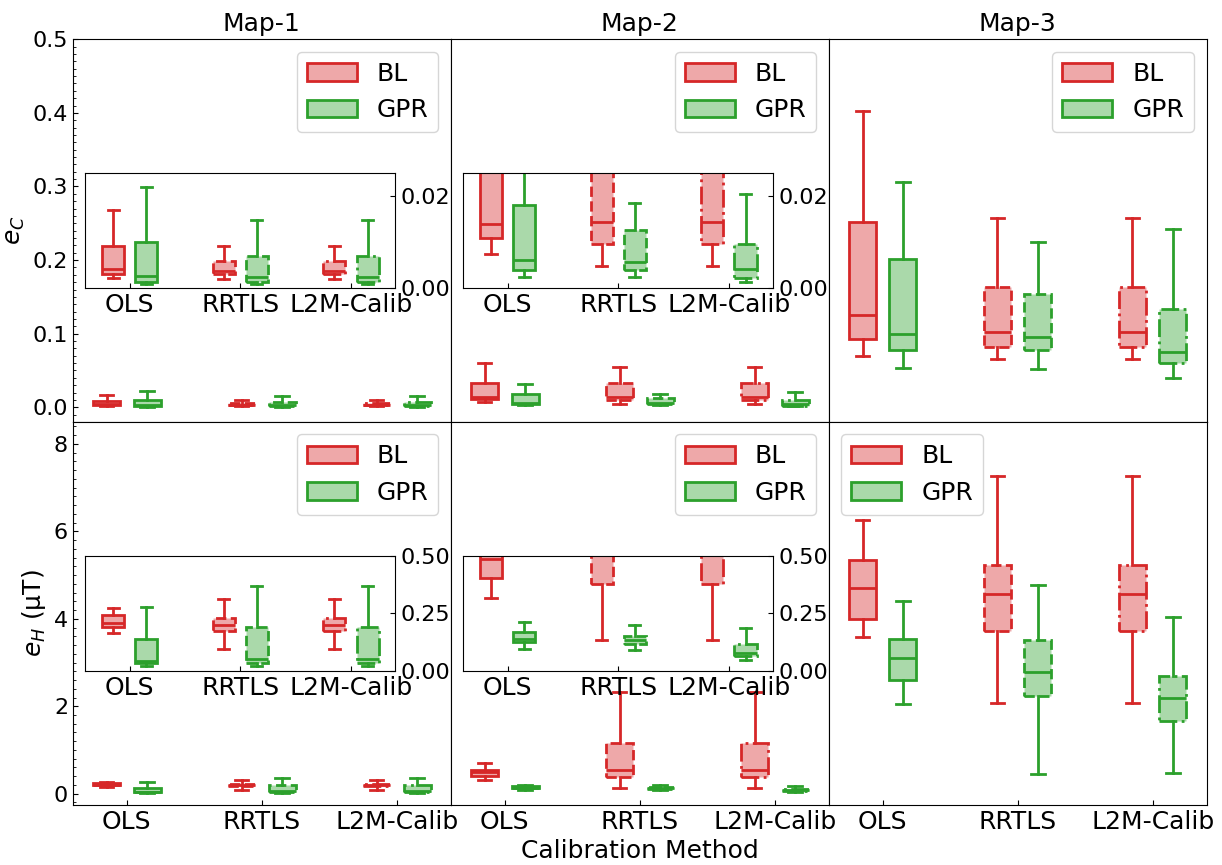}
			\centering
		}
	}
	\caption{Demonstration of the calibration results under the usage of different magnetic map interpolation methods and calibration strategies. (a) Interpolation result of magnetic map under different fingerprint density. Different colors represent different magnetic total intensity; (b) Calibration results using different mangetic map in (a) with different calibration strategies - OLS, RRTLS and \textit{L2M-Calib} (ours). The results are shown in errorness compared with ground truth value.}
	\label{fig: map_interpolation}
	\vspace{-1.5em}
\end{figure*}
We study the effect of the interpolation method and the weighting mechanism in w-RRTLS of \textit{L2M-Calib}. Magnetic maps are generated using two interpolation schemes, Bi-Linear and sliding Gaussian Process Regression (s-GPR), at three fingerprint densities shown in Fig.~\ref{fig: map_interpolation}(a). These maps are used as references in calibration with OLS, RRTLS, and \textit{L2M-Calib}. For each setting, calibration is repeated 50 times using randomly sampled intrinsic parameters and Gaussian noise with $\sigma = 0.5\mu T$.

The results Fig.~\ref{fig: map_interpolation}(b) show that even when the interpolation method has limited impact at high fingerprint densities, s-GPR outperforms Bi-Linear interpolation by reducing mean bias errors by over $0.2\mu T$. When the fingerprint is sparse, this improvement is much more significant. Moreover, the use of weights in \textit{L2M-Calib} effectively suppresses outliers and consistently improves calibration accuracy across all fingerprint densities compared with using RRTLS directly.

\subsection{Qualitative Analysis}
To demonstrate practical utility, we test the method on real-world platforms. First, two distinct magnetic fingerprints are collected using the ideal (undistorted) system, and used to build a calibration map and a validation map. Then, distortion is introduced by attaching ferromagnetic materials (steel bar and block) on the plastic trolley and a more distortion-prone Bunker AGV platform. The calibration is conducted on the distorted system using the calibration map, and performance is verified on the validation map. Success is measured by the mean and standard deviation of the difference between the calibrated readings and the validation map. 
\begin{figure*}[!t]
	\centering{
		\subfloat[Calibration result validation for trolley scenario]{
			\includegraphics[width=0.49\linewidth]{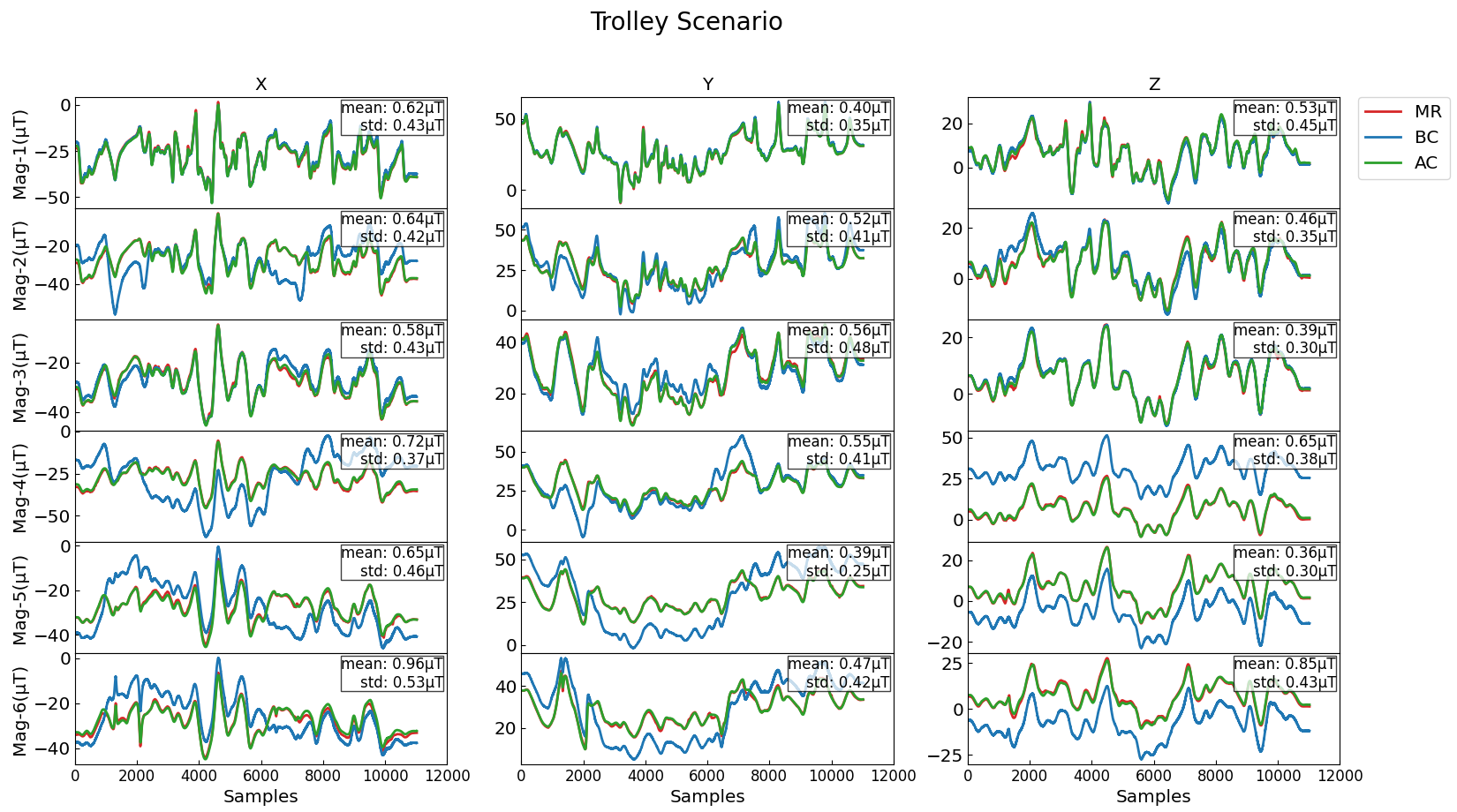}
			\centering
		}
		\subfloat[Calibration result validation for AGV scenario]{
			\includegraphics[width=0.49\linewidth]{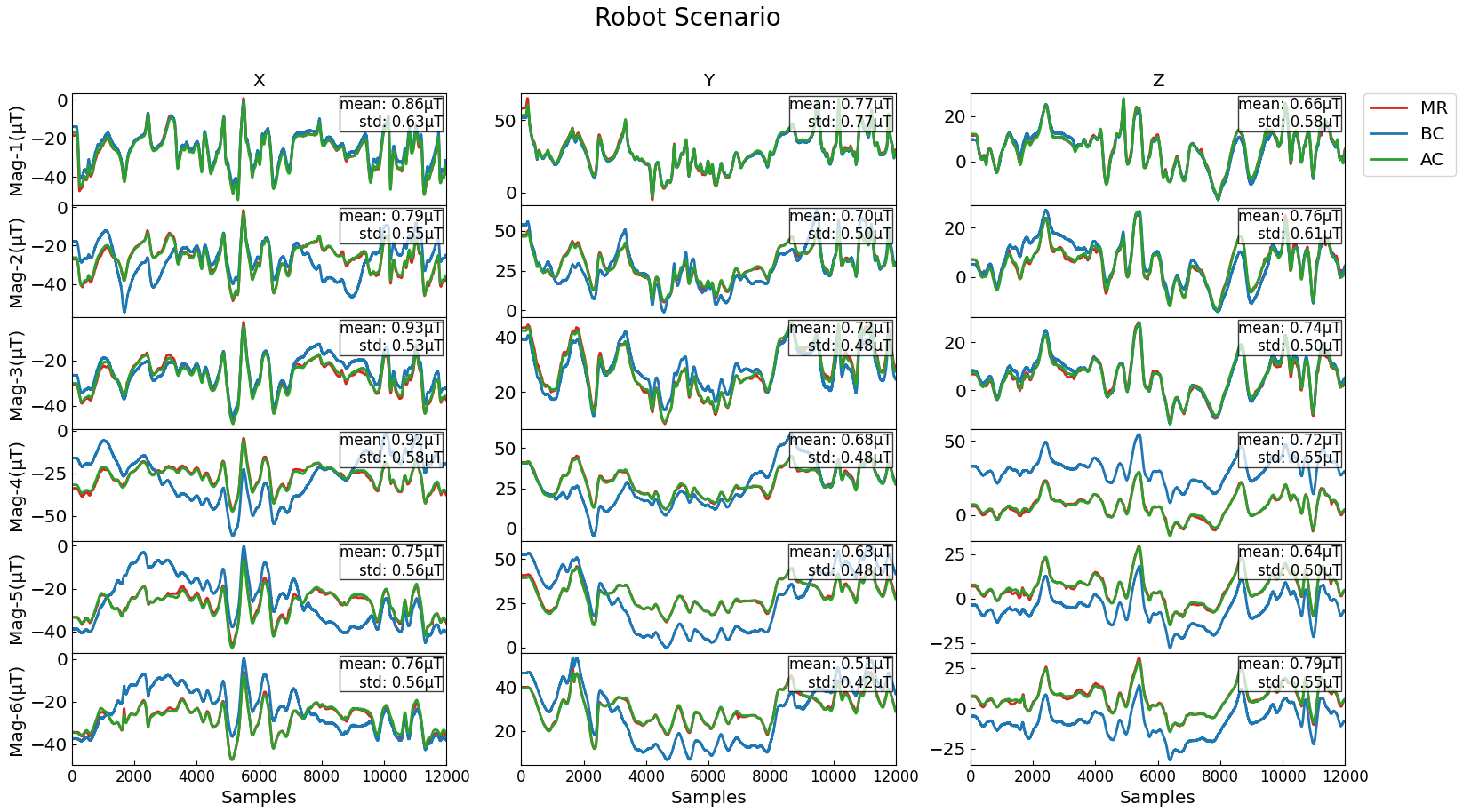}
			\centering
		}
	}
	\caption{Calibration results of both trolley scenario and Bunker scenario by comparing the distorted magnetic sensors' readings (BC), recovered magnetic sensors' readings (AC) with the reference validation map's reading (MR). }
	\label{fig: real_world_calibration}
	\vspace{-1.5em}
\end{figure*}

The final calibration results are shown in Fig.~\ref{fig: real_world_calibration} for both trolley and Bunker AGV platforms. Each column presents X, Y, and Z magnetic readings while each row is one of six sensors to calibrate. The red line (MR) indicates the reference validation map, the blue line (BC) shows the uncalibrated readings, and the green line (AC) shows the calibrated output. After calibration, the calibration output readings closely align with the reference values, regardless of platform or direction, unlike the uncalibrated readings. The mean error is below $1 \mu T$ on all axes with low variance, demonstrating consistent and effective calibration across platforms.

\section{Conclusion}\label{sec: section5}
In this paper, we have presented a novel calibration method, \textit{L2M-Calib}, for magnetic-LiDAR fused systems deployed on autonomous ground vehicles. The proposed approach jointly estimates the extrinsic transformation between the magnetic and LiDAR frames, as well as the intrinsic distortion parameters of the magnetic sensor just in one-key. By formulating the problem as a two-step optimization, we achieve robust and accurate calibration even under challenging conditions using AGV. Extensive simulation and real-world experiments validate the effectiveness, accuracy, and robustness of our approach across different sensor configurations and motion odometry. These results demonstrate the proposed method’s suitability for diverse deployment scenarios in magnetic-LiDAR fused autonomous navigation and perception systems.

\bibliographystyle{ieeetr}
\bibliography{reference}

\end{document}